\documentclass[11pt]{article}
\usepackage{named}

\usepackage{mathptmx}
\usepackage{amsmath}
\usepackage{amssymb}
\usepackage{listings}
\usepackage{verbatim}
\usepackage{float}
\def\ii#1{\hbox{\it #1\/}}

\def\beq{\begin{equation}}
\def\eeq#1{\label{#1}\end{equation}}
\def\ba{\begin{array}}
\def\ea{\end{array}}

\begin{document}
\bibliographystyle{named}

\title{\bf Achievements in Answer Set Programming}

\author{Vladimir Lifschitz}

\maketitle

\begin{abstract}
This paper describes an approach to the methodology of answer set programming 
that can facilitate the design of encodings that are easy to understand 
and provably correct.  Under this approach, after appending a rule or a small
group of rules to the emerging program we include a comment that states what
has been ``achieved'' so far.   This strategy allows us to set out our
understanding of the design of the program by describing the roles of
small parts of the program in a mathematically precise way.
\end{abstract}

\section{Introduction}

This paper describes an approach to the methodology of answer set
programming \cite{mar99,nie99} that can facilitate the design of encodings
that are easy to understand and provably correct.  Under this approach,
after appending a rule or a small group of rules to the emerging program,
the programmer would include a comment that states what has been
``achieved'' so far, in a certain precise sense.

Consider, for instance, the following solution to the 8 queens problem,
adapted from \cite[Section~3.2]{geb12}.\footnote{A concise introduction
to ASP can be found in Chapter~1 of that book.  Examples of
programs in this paper are written in the input language of the grounder
{\sc gringo}, Version~5.}
\medskip

\begin{lstlisting}
% Program 8Queens

row(1..8).
col(1..8).                                     
8 { queen(I,J) : col(I), row(J) } 8.
:- queen(I,J), queen(I,JJ), J!=JJ.                          
:- queen(I,J), queen(II,J), I!=II.               
:- queen(I,J), queen(II,JJ), (I,J)!=(II,JJ), |I-II|=|J-JJ|. 
\end{lstlisting}

\medskip\noindent
The first rule of \ii{8Queens} (Line~3), viewed as a one-rule program, has a
unique stable model~$S$, which satisfies the following condition:
\beq
\hbox{A ground atom of the form $\ii{row}(i)$ belongs to~$S$
iff~$i\in\{1,\dots,8\}$.}
\eeq{a1}
Condition~(\ref{a1}) holds also if~$S$ is the stable model of
the first two rules of this program.  And it holds if~$S$ is any stable
model of the first three rules, and so on, for all 6 ``prefixes'' (initial
segments) of the program.  This is what we mean by achievement: once the
programmer declares that a property ``has been achieved,'' he is committed
to maintaining this property of stable models until the program is completed.

After writing the second rule (Line~4), the programmer can claim that
something else has been achieved:
\beq
\hbox{A ground atom of the form $\ii{col}(j)$ belongs to~$S$
iff~$j\in\{1,\dots,8\}$.}
\eeq{a2}
This condition holds if~$S$ is a stable model of any prefix of
the program that includes the first two rules.

Additional properties achieved by adding the third rule can be expressed as
follows:
\beq
\ba l
\hbox{Set~$S$ contains exactly 8 ground atoms of the form
$\ii{queen}(i,j)$.}\\
\hbox{For each of these atoms, $i,j\in\{1,\dots,8\}$.}
\ea
\eeq{a3}

If a program is written in this manner then every achievement
documented in the process of writing it describes a property shared by
all stable models of the entire program.  In some cases this list of
achievements can serve as the skeleton of a proof of its correctness,
in the spirit of Edsger Dijkstra's advice:
\begin{quote}
    $\dots$ one should not first make the program and then prove its
    correctness, because then the requirement of providing the
    proof would only increase the poor programmer's burden. On the
    contrary: the programmer should let correctness proof and
    program grow hand in hand \cite{dij72}.
\end{quote}
Recording important achievements in the process of writing
an ASP program may be similar to recording important loop invariants
in procedural programming: it does not ensure the correctness
of the program but helps the programmer move toward the goal of
proving correctness.

A preliminary report on this project was presented at the 2016 Workshop
on Answer Set Programming and Other Computing Paradigms.  This is a
corrected version of the paper that appeared in {\sl Theory and Practice
of Logic Programming}.

\section{Programs, Prefixes, and Achievements}\label{sec:ppa}

In this paper, by an (ordered)
program we understand a list of rules $R_1,\dots,R_n$
($n\geq 1$) in the input language of an answer set solver, such as {\sc
clingo} \cite{gringomanual} or {\sc dlv}~\cite{eit98}.
The order of rules is supposed to
reflect the order in which the programmer writes them in the process
of creating the program.  It does not affect the semantics of the program,
but it is essential for understanding the process of programming.

We restrict attention to programs without classical negation.  (This
limitation is discussed in the conclusion.)  Stable
models of a program without classical negation are sets of ground atoms
that contain no arithmetic operations, intervals, or pools
\cite[Sections~3.1.7, 3.1.9, 3.1.10]{gringomanual}.  Such ground atoms will
be called {\sl precomputed}.\footnote{This terminology follows
Gebser et al.~[\citeyear{geb15}], where ``precomputed terms'' are
defined.  Calimeri et al.~[\citeyear{aspcore2}] talk about elements of
the ``Herbrand universe'' of a program in the same sense.}
An {\sl interpretation} is a set of precomputed atoms.

The {\sl $k$-th prefix} of a program $R_1,\dots,R_n$, where $1\leq k\leq n$,
is the program $R_1,\dots,R_k$.  We will express that a program~$\Gamma$ is
a prefix of a program~$\Pi$ by writing $\Gamma\leq\Pi$.  The relation
$\leq$ is a total order on the set of prefixes of a program.

An {\sl achievement} of a prefix~$\Gamma$ of~$\Pi$ is a property of sets
of interpretations that holds for all stable models of all programs~$\Delta$
such that $\Gamma\leq\Delta\leq\Pi$.
For example,~(\ref{a1}) is an achievement of the first prefix of program
\ii{8Queens\/};~(\ref{a2}) is an achievement of its second
prefix; and~(\ref{a3}) is an achievement of its third prefix.
Conditions~(\ref{a1}) and~(\ref{a2}), and the conjunction of
conditions \hbox{(\ref{a1})--(\ref{a3})}, are achievements of the third
prefix
as well.  Any condition that holds for all sets of interpretations is
trivially an achievement of any prefix of any program.

The following three conditions are achievements of the last
three prefixes of \ii{8Queens\/}:
\beq
\ba l
\hbox{Each column of the $8\times 8$ chessboard includes at most one
square~$(i,j)$}\;\\
\hbox{such that the atom~$\ii{queen}(i,j)$ belongs to $S$.}\;
\ea
\eeq{a4}
\beq
\ba l
\hbox{Each row of the $8\times 8$ chessboard includes at most one
square~$(i,j)$}
\qquad\\
\hbox{such that the atom~$\ii{queen}(i,j)$ belongs to $S$.}
\qquad
\ea
\eeq{a5}
\beq
\ba l
\hbox{Each diagonal of the $8\times 8$ chessboard includes at most one
square~$(i,j)$}\\
\hbox{such that the atom~$\ii{queen}(i,j)$ belongs to $S$.}
\ea
\eeq{a6}
Thus every stable model~$S$ of \ii{8Queens} satisfies all
conditions \hbox{(\ref{a1})--(\ref{a6})}.

\section{Programs with Input}\label{sec:pip}

In some programs, constants are used as placeholders for values provided by
the user \cite[Section~3.1.15]{gringomanual}.  For example, the constant~$n$
is used as a placeholder for an arbitrary positive integer in the
following more general version of \ii{8Queens}:
\medskip

\begin{lstlisting}
% Program NQueens

row(1..n).
col(1..n).                                     
n { queen(I,J) : col(I), row(J) } n.
:- queen(I,J), queen(I,JJ), J!=JJ.                          
:- queen(I,J), queen(II,J), I!=II.               
:- queen(I,J), queen(II,JJ), (I,J)!=(II,JJ), |I-II|=|J-JJ|. 
\end{lstlisting}

The value of a placeholder is one kind of input that an answer
set solver may expect in addition to the rules of the program.
A definition of an ``extensional predicate'' occurring in the bodies
of rules is another kind.  Consider, for example, the following encoding
of Hamiltonian cycles, adapted from \cite[Section 3.3]{geb12}:
\medskip

\begin{lstlisting}
% Program Hamiltonian

1 {in(X,Y) : edge(X,Y) } 1 :- vertex(X).
1 {in(X,Y) : edge(X,Y) } 1 :- vertex(Y).
reached(X) :- in(v0,X).
reached(Y) :- reached(X), in(X,Y).
:- not reached(X), vertex(X).
\end{lstlisting}

\medskip\noindent
It needs to be supplemented by definitions of
the predicate $\ii{vertex}/1$, representing the set of vertices of a finite
digraph $G$; of
the predicate $\ii{edge}/2$, representing the set of edges of $G$;
and of the placeholder $v0$, which is a vertex of $G$.

In general, we understand an {\sl input} as a function $\bf i$ defined on a finite set
consisting of predicate symbols  and symbolic constants such that
\begin{itemize}
\item if~$p/m$ is a predicate symbol in the domain of~$\bf i$ then
${\bf i}(p/m)$ is a finite set of $m$-tuples of precomputed terms;
\item
if~$c$ is a symbolic constant in the domain of~$\bf i$ then ${\bf i}(c)$
is a precomputed term.\footnote{Programs with input in the sense of this
definition
are similar to lp-functions in the sense of Gelfond and Przymusinska
[\citeyear{DBLP:journals/ijseke/GelfondP96}].  The description of
an lp-function specifies not only its input, but also its output; on the
other hand, the input of an lp-function includes predicates only, not
placeholders.}
\end{itemize}
The result of {\sl enriching} a program~$\Pi$ by an input~$\bf i$, denoted by
$\Pi\diamond {\bf i}$, is the program consisting of
\begin{itemize}
\item
the facts $p({\bf t})$ for all predicate symbols~$p/m$ in the domain of~$\bf i$
and all tuples ${\bf t}$ in ${\bf i}(p/m)$, followed by
\item
the rules obtained from the rules of~$\Pi$ by substituting the terms ${\bf i}(c)$ for
all occurrences of symbolic constants~$c$ in the domain of~$\bf i$.
\end{itemize}
The stable models of $\Pi\diamond {\bf i}$ will be called the {\sl stable models of~$\Pi$
for input~$\bf i$}.

For example, if {\bf i} is the input that maps $n$ to 8 then
$\ii{NQueens}\diamond {\bf i}$ is the program \ii{8Queens}.  If {\bf i} is
the input that maps $\ii{vertex}/1$ to $\{a,b\}$, $\ii{edge}/2$ to
$\{(a,b),(b,a)\}$, and \ii{v0} to~$a$, then $\ii{Hamiltonian}\diamond{\bf i}$ is
the program consisting of the facts
\begin{verbatim}
               vertex(a).  vertex(b).  edge(a,b).  edge(b,a).
\end{verbatim}
followed by the rules of \ii{Hamiltonian} with \ii{v0} replaced by $a$.

We will now extend the definition of an achievement to programs for which some
inputs are designated as ``valid.''  For example, we can say that an
input {\bf i} is considered valid for
the program \ii{NQueens} if its domain includes only one object---the symbolic
constant $n$---and if~${\bf i}(n)$ is a positive integer.  An input~$\bf i$ is considered valid for the
program \ii{Hamiltonian} if (a)~its domain consists of three objects---the
predicate symbols~$\ii{vertex}/1$ and~$\ii{edge}/2$ and the symbolic
constant~\ii{v0}; (b)~${\bf i}(\ii{edge}/2)$ is the set of edges of a digraph
with the set of vertices~${\bf i}(\ii{vertex}/1)$; and (c)~~${\bf i}(\ii{v0})$
is a vertex of that graph.

In this setting, an achievement is
a relation between valid inputs and sets of interpretations.  Such a
relation~$A$ is an {\sl achievement} of a prefix~$\Gamma$ of~$\Pi$
if, for every program~$\Delta$ such that $\Gamma\leq\Delta\leq\Pi$,
$A({\bf i},S)$ holds for every valid input~{\bf i} and every stable model~$S$
of~$\Delta$ for input~{\bf i}.  In particular, if~$A$ is an achievement of the
entire program~$\Pi$ then $A({\bf i},S)$ holds for every valid input~{\bf i}
and every stable model~$S$ of~$\Pi$ for that input.

For example, the sentence
\beq
\hbox{A ground atom of the form $\ii{row}(i)$ belongs to~$S$
iff~$i\in\{1,\dots,{\bf i}(n)\}$}
\eeq{a1n}
expresses a relation between~{\bf i} and~$S$ that is an achievement of the
first prefix of \ii{NQueens}.  It is obtained from
condition~(\ref{a1}) by replacing 8 with~${\bf i}(n)$, and achievements of the
other prefixes of that program can be obtained in a similar way from
conditions~(\ref{a2})--(\ref{a6}).

The following two conditions are achievements of the first two prefixes
of \ii{Hamiltonian\/}:
\beq
\ba l
\!\!\!\!\hbox{Every pair~$(x,y)$ such that the atom $\ii{in}(x,y)$ belongs
to~$S$ is an edge of the }\\
\!\!\!\!\hbox{digraph~$G$ with the vertices ${\bf i}(\ii{vertex}/1)$ and edges
${\bf i}(\ii{edge}/2)$; for every}\\
\!\!\!\!\hbox{vertex~$x$ of~$G$ there is a unique~$y$ such
that the atom~$\ii{in}(x,y)$ belongs to~$S$.}
\ea
\eeq{h1}
\beq
\ba l
\,\hbox{For every vertex~$y$ of~$G$ there is a unique~$x$ such that the
atom $\ii{in}(x,y)$ belongs to~$S$.}
\ea
\eeq{h2}
Nothing interesting has been achieved by adding the third rule, but
the following condition is an achievement of the fourth prefix of the
program:
\beq
\ba l
\hbox{The set of symbols~$x$ such that the atom $\ii{reached}(x)$
belongs to~$S$ consists}\\
\hbox{of the vertices~$x$ of~$G$ for which there exists a
walk~$v_0,\dots,v_n$ such that}\\
\hbox{ $n\geq 1$, $v_0=v0$, $v_n=x$, and every atom of the form $\ii{in}(v_i,v_{i+1})$
belongs to~$S$.}
\ea
\eeq{h4}
Finally, here is an achievement of the entire program:
\beq
\hbox{For every vertex~$x$ of~$G$, the atom $\ii{reached}(x)$ belongs
to~$S$.}
\eeq{h5}

\section{Records of Achievement}\label{sec:ra}

A {\sl record of achievement} for a program~$\Pi$ is described by
assigning an achievement~$A_\Gamma$ to each~$\Gamma$ in the record's {\sl
domain}, which is
a set of prefixes of~$\Pi$ that includes the entire program~$\Pi$.
We will represent a record of achievement by the listing of the program
with comments describing the achievements~$A_\Gamma$ placed vafter all
prefixes~$\Gamma$ in the record's domain.

For example, here is the program \ii{NQueens} with a record of achievement:

\medskip
\begin{lstlisting}
% Program NQueens, with a record of achievement

% input: positive integer n (the size of the board).

% A square on the board is represented as a pair, column
% number and row number, both from the set {1,..,n}.

row(1..n).
% achieved: row/1 = {1,...,n}.

col(1..n).                                     
% achieved: col/1 = {1,...,n}.

n { queen(I,J) : col(I), row(J) } n.
% achieved: Set queen/2 consists of n squares.

:- queen(I,J), queen(I,JJ), J!=JJ.                          
% achieved: Each column includes at most one square from queen/2.

:- queen(I,J), queen(II,J), I!=II.               
% achieved: Each row includes at most one square from queen/2.

:- queen(I,J), queen(II,JJ), (I,J)!=(II,JJ), |I-II|=|J-JJ|. 
% achieved: Each diagonal includes at most one square from
%           queen/2.
\end{lstlisting}

\medskip\noindent
The domain of this record achievement is the set of all prefixes of the
program.  The comment in Line~3 shows
which inputs for the program are considered valid.
The comment in Line~9 is a concise reformulation of condition~(\ref{a1n}).
It uses $\ii{row}/1$ as shorthand for ``the set of precomputed terms~$i$
such that the atom $\ii{row}(i)$ belongs to~$S$.''  In the other comments,
$\ii{col}/1$ and $\ii{queen}/2$ are understood in a similar way.

Program \ii{Hamiltonian} with a record of achievement below
uses another useful convention: in Lines~11 and 12,
we understand~$X$ and~$Y$ as metavariables for precomputed terms.  The
comment in those lines is a reformulation of condition~(\ref{h2}).
\medskip

\begin{lstlisting}
% Program Hamiltonian, with a record of achievement

% input: the set vertex/1 of vertices of a finite digraph G;
%        the set edge/2 of edges of G; a vertex v0 of G.

1 {in(X,Y) : edge(X,Y) } 1 :- vertex(X).
% achieved: Set in/2 is a subset of edge/2; for every vertex
%           X of G there is a unique Y such that in(X,Y).

1 {in(X,Y) : edge(X,Y) } 1 :- vertex(Y).
% achieved: For every vertex Y of G there is a unique X
%           such that in(X,Y).

reached(X) :- in(v0,X).
reached(Y) :- reached(X), in(X,Y).
% achieved: Set reached/1 consists of the vertices that are
%           reachable from v0 by a path of non-zero length
%           in the subgraph of G with the set of edges in/2.

:- not reached(X), vertex(X).
% achieved: reached/1 = vertex/1.
\end{lstlisting}

\section{Completeness}

The record of achievement for program \ii{NQueens} in Section~\ref{sec:ra}
shows that every stable model~$S$ of that program for the input $n=8$
satisfies conditions~(\ref{a1})--(\ref{a6}).  The converse is not true:
some sets of precomputed atoms satisfying all these conditions are not
stable models.  This is clear from the fact that these conditions
say nothing about precomputed atoms formed using predicate symbols
other than
\beq
\ii{row}/1,\; \ii{col}/1,\; \ii{queen}/2.
\eeq{voc}
Adding such
``irrelevant'' atoms to a stable model of the program would not invalidate
properties \hbox{(\ref{a1})--(\ref{a6})}.  It is true, however, that if every
atom in~$S$ contains one of the symbols~(\ref{voc}) then~$S$ has all
these properties if {\sl and only if} it is a stable model for $n=8$.

This observation leads us to the following definitions.  The {\sl vocabulary}
of a program~$\Pi$ is the set of all precomputed atoms $p(t_1,\dots,t_n)$
such that the predicate symbol $p/n$ occurs in~$\Pi$.  A record of
achievement $\Gamma\mapsto A_\Gamma$ for~$\Pi$ is {\sl complete}
if, for every valid input~{\bf i}, each subset~$S$ of the vocabulary
of~$\Pi\diamond{\bf i}$ that satisfies conditions $A_\Gamma({\bf i},S)$ for
all prefixes $\Gamma$ in the record's domain is a stable model of~$\Pi$ for
input~{\bf i}.  The converse---every stable model~$S$ of~$\Pi$  for
input~{\bf i} is a subset of the vocabulary satisfying these
conditions---is true for any record of achievement.  Thus
achievements in a complete record provide a complete characterization of
the class of stable models of the program.

For example, the record of achievement for \ii{NQueens}
% and \ii{Hamiltonian}
in Section~\ref{sec:ra} is complete.  This property can be lost if we
make the achievements in that record weaker.  For instance, if we replace
 ``consists of~$n$ squares'' in Line 15 by ``consists of at most~$n$
squares'' then we will not eliminate the possibility that
$\ii{queen}/2$ is empty.

The completeness property holds not only for the entire record
of achievement for \ii{NQueens} above, but also for its initial segments.  For
instance, a set~$S$ of precomputed atoms formed using $\ii{row}/1$
satisfies~(\ref{a1}) only if it is a stable model of the first rule of
the program for $n=8$.  A set of precomputed atoms formed using
$\ii{row}/1$, $\ii{col}/1$ satisfies~(\ref{a1}) and~(\ref{a2}) only if it
is a stable model of the first two rules, and so forth.

The record of achievement for \ii{Hamiltonian} in Section~\ref{sec:ra} is
complete also, and so are its initial segments.  For example, a set~$S$ of
precomputed atoms formed using the predicate symbols $\ii{vertex}/1$,
$\ii{edge}/2$, and $\ii{in}/2$ satisfies condition~(\ref{h1}) if and only if
it is a stable model of the first rule of the program for input~{\bf i}.

\section{Achievement-Based Answer Set Programming}
\label{sec:abasp}

Both records of achievement given as examples in Section~\ref{sec:ra} are
not only complete but
also detailed, in the sense that they include achievements for almost all
prefixes of the programs.  The only rule in these programs that is not
followed by an achievement comment is the first rule in the recursive
definition of $\ii{reached}$. The role of that rule cannot be properly
explained unless we treat it as part of the definition.

Developing an ASP program along with a complete and detailed record of
achievement can be called ``achievement-based'' answer set programming.
This strategy allows us to set out our understanding of the design of
the program by describing the roles of individual rules, or small groups
of rules, in a mathematically precise way.

One of the advantages of this approach is that comments explaining what
is achieved by a group of rules at the beginning of a program help us
start testing and debugging it at an early stage, when only a part of the
program has been written.  To this end, we can run an answer set solver to
find stable models of the prefix that has has been already
written and check that they satisfy the conditions in the available
``achieved'' comments.  A mismatch would indicate that there is a bug in
the rules of the program written so far, or perhaps that the programmer's
intentions have not been properly documented in the recorded
achievements.

A complete record of achievement is particularly valuable when it is closely
related to the program's specification, because from the completeness of
such a record we may be able to conclude that the program is correct.
For instance, from the completeness of the record of achievement of
\ii{NQueens} in Section~\ref{sec:ra} we can conclude that the stable models
of that program are in a one-to-one correspondence with solutions to the
$n$ queens problem.

To further illustrate the idea of achievement-based ASP, we present below
three ``real life'' ASP programs accompanied by complete, detailed
records of achievement.  The first of them,
program \ii{SCA} \cite[Figure~1]{bra12}, generates sequence covering
arrays\footnote{A sequence covering array of strength~$t$ is an array
of permutations of symbols such that every ordering of any~$t$ symbols
appears as a subsequence of at least one row.} of strength~3.
Our version of \ii{SCA} is slightly different from the original program: the
constraint in Line~19 here replaces the pair of rules
\begin{verbatim}
               hb(N,X,Z) :- hb(N,X,Y), hb(N,Y,Z).
               :- hb(N,X,X).
\end{verbatim}
The reason why we chose to make this change is that the first of the two
rules above may temporarily destroy the irreflexivity of the relation
of $\ii{hb}_N$ that was true at the previous step; that property is
restored by the second rule.  That is not in the spirit of the
achievement-based approach, which emphasizes the gradual accumulation of
properties that we would like to see in the complete program.
\medskip

\begin{lstlisting}
% Program SCA

% input: the number s of symbols 1,...,s; the number n of
%        rows 1,...,n.

sym(1..s).
% achieved: sym/1 = {1,...,s}.

row(1..n).
% achieved: row/1 = {1,...,n}.

1 {hb(N,X,Y); hb(N,Y,X)} 1 :- row(N), sym(X), sym(Y), X!=Y.
% For every row N, let hb_N be the binary relation on sym/1
% defined by the condition: X hb_N Y iff hb(N,X,Y).
% achieved: each relation hb_N is irreflexive; each pair of
%           distinct symbols satisfies either X hb_N Y
%           or Y hb_N X.

:- hb(N,X,Y), hb(N,Y,Z), not hb(N,X,Z).
% achieved: each relation hb_N is transitive.

covered(X,Y,Z) :- hb(N,X,Y), hb(N,Y,Z).
% For every row N and every symbol X, by M_{N,X} we denote
% the symbol that is the X-th smallest w.r.t. hb_N.
% achieved: for any symbols X, Y, Z, covered(X,Y,Z) iff,
%           for some row N, (X,Y,Z) is a subsequence of
%           (M_{N,1},...,M_{N,s}).
            
:- not covered(X,Y,Z), sym(X), sym(Y), sym(Z),
   X!=Y, Y!=Z, X!=Z.
% achieved: covered(X,Y,Z) for any pairwise distinct symbols
%           X, Y, Z.
\end{lstlisting}

The other two examples are program \ii{Borda}, adapted from
\cite[Encoding~1]{cha15}, which encodes
the Borda rule for determining the winner in an election with several
candidates,\footnote{Each voter ranks the list of candidates in order of
preference. The candidate ranked last gets zero points; next to last gets
one point, and so on.  The candidate with the most points is the winner.}
\pagebreak

$ $
\vskip 3cm
\begin{lstlisting}
% Program Borda

% input: the number m of candidates 1,...,m in an election
%        E; the set p/3 of triples (P,Pos,C) such that, for
%        a fixed ordering pr_1,...,pr_l of the distinct
%        preference relations in the profile of E, candidate
%        C is at position Pos in relation pr_P; the set
%        votecount/2 of pairs (P,VC) such that relation pr_P
%        occurs VC times in the profile of E.

candidate(1..m).
% achieved: candidate/1 = {1,...,m}.

posScore(P,C,X*VC) :- p(P,Pos,C), X=m-Pos, votecount(P,VC).
% achieved: posScore(P,C,S) iff the voters who chose
%           relation pr_P in election E contributed S points
%           to candidate C under the Borda rule.

score(C,N) :- candidate(C), N=#sum{S,P:posScore(P,C,S)}.
% achieved: score(C,N) iff candidate C earned N points in
%           election E under the Borda rule.

winner(C) :- score(C,M), M=#max{S:score(_,S)}.
% achieved: winner(C) iff the number of points earned by
%           candidate C in election E is maximal among all
%           candidates.
\end{lstlisting}

\pagebreak\noindent
and program \ii{OBT},
adapted from \cite[Section~1]{bro07}, which encodes ordered binary
trees.\footnote{An ordered binary tree is a
rooted binary tree with the leaves $0,\dots,k$ and internal vertices
$k+1,\dots,2k$ such that (i)~every internal vertex is greater than its
children, and (ii) for any two internal vertices $x$ and $x_1$,
$x>x_1$ iff the maximum of the children of~$x$ is grater than the maximum of
the children of~$x_1$.}

\begin{lstlisting}
% Program OBT

% input: positive integer k.

leaf(0..k).
% achieved: leaf/1 = {0,...,k}.

vertex(0..2*k).
% achieved: vertex/1 = {0,...,2k}.

internal(X) :- vertex(X), not leaf(X).
% achieved: internal/1 = {k+1,...,2k}.

2 {edge(X,Y) : vertex(Y), X>Y} 2 :- internal(X).
% Let G be the digraph with the vertices vertex/1 and the
% edges edge/2.
% achieved: for every edge (X,Y) of G, X>Y; the out-degree
%           of a vertex X in G is 2 if internal(X), and 0
%           if leaf(X).

reachable(X,Y) :- edge(X,Y).
reachable(X,Y) :- edge(X,Z), reachable(Z,Y).
% achieved: reachable(X,Y) iff Y is reachable from X in G
%           by a path of non-zero length.

:- vertex(X), X!=2*k, not reachable(2*k,X).
% achieved: every vertex of G other than 2k is reachable
%           from 2k by a path of non-zero length.

:- reachable(X,X), vertex(X).
% achieved: G is acyclic.

max_child(X,Y) :- edge(X,Y), edge(X,Y1), Y > Y1.
% achieved: max_child(X,Y) iff Y is the largest child of X
%           in G.

Y<Y1 :- max_child(X,Y), max_child(X1,Y1), Y>Y1, X<X1.
% achieved: for any vertices X, X1 of G such that X<X1, the
%           largest child of X is smaller than the largest
%           child of X1.
\end{lstlisting}
\pagebreak

\section{Achievements in Teaching}

The achievement-based approach was emphasized in a class on answer set
programming taught recently to a group of over 50 undergraduates at
the University of Texas at Austin.  The idea of an achievement was explained
more informally than in this paper, but many examples were given.
In most solutions to programming assignments submitted for grading,
students attempted to imitate the instructor's use of ``input'' and
``achieved'' comments, even though they were not instructed to do that.
The degree of their success depended, of course, on their previous
exposure to logic and mathematics.  When ASP programs written by students
were discussed in class, the instructor emphasized the difference between
the correctness of the program on the one hand, and the clarity and
correctness of ``input'' and ``achieved'' comments on the other.

Comments of these kinds can be used in exercises and test
problems.  In one case, students were shown an ``incomplete listing'' of a
graph coloring program:

\medskip
\begin{lstlisting}
% Color the vertices of a graph so that no two adjacent
% vertices share the same color.

% input: set vertex/1 of vertices of a graph G;
%        set edge/2 of edges of G; set color/1 of colors.

1 {color(X,C) : color(C)} 1 :- vertex(X).
% achieved: for every vertex X there is a unique color C
%           such that color(X,C).

_________________________________________________________
% achieved: no two adjacent vertices share the same color.

#show color/2.
\end{lstlisting}

\medskip\noindent
The question was, ``What rule would you place in Line~11?''  On
another occasion, students were asked to write a one-rule program for
which the following comments would be appropriate:

\medskip
\begin{lstlisting}
% Calculate the number of classes taught today on each of
% the seven floors of the computer science building.

% input: set where/2 of all pairs (C,I) such that class C
%        is taught on the I-th floor.

_________________________________________________________
% achieved: howmany(I,N) iff the number of classes taught
%           on the I-th floor is N.

#show howmany/2.
\end{lstlisting}
\section{Conclusion}

In achievement-based ASP, we start writing a program by describing its
inputs.  Then, after every rule or small group of rules, we include a
comment describing what has been achieved.  Collectively these comments
represent a complete, detailed record of achievement.

As we are adding rules to an emerging ASP program, we deal at every step
with a single executable piece of code, unlike the non-executable
pseudo-code formed in the process of stepwise refinement of a
procedural program, and unlike a collection of executable subroutines
formed in the process of bottom-up design.
In the process, we think of prefixes of the emerging program as if they were
complete programs.  We describe their stable models in a way that relates
them to the stable models of the final product.

The programs discussed in this paper do not use classical negation
\cite{gel90}.
In the presence of classical negation, answer sets consist of
``precomputed classical \hbox{literals}''
---precomputed atoms and classical
negations of such atoms.  Extending the definition of a complete record of
achievement to such programs is straightforward.  On the other hand,
many programs with classical negation contain defaults
\cite[Chap.~5]{gel14}, such as the closed world assumption and the
commonsense law of inertia, and the achievement-based
approach may be not so useful in application to programs containing defaults.
A default does not ``achieve'' anything in the technical sense of
Section~\ref{sec:ppa}.

When ASP is used for representing dynamic domains, a very different
methodology can be recommended: first describe the domain in an action
description language, and then translate its causal laws into answer
set programming \cite{gel93a}, \cite{lif99b}, \cite{gel14}.

According to Gebser et al.~[\citeyear{geb12}],
\begin{quote}
[t]he basic approach to writing encodings in ASP follows a {\sl
generate-and-test} methodology, also referred to as
{\sl guess-and-check}$\dots$ A ``generating'' part is meant to
non-deterministically provide solution
candidates, while a ``testing'' part eliminates candidates violating some
requirements$\dots$ Both parts are usually amended by ``defining'' parts
providing auxiliary concepts.
\end{quote}
Most programs discussed in this paper are designed in accordance
with this basic approach.\footnote{Program \ii{Borda} is
an exception---it has no generating part and no testing part.  Also, it
is not clear whether the designers of \ii{Hamiltonian}
intended the second rule for the generating part
or for the testing part.  (The second rule is syntactically similar to the
first, which is definitely a generate rule.  On the other hand,
adding the second rule does not really generate new
solution candidates; it eliminates some of the candidates generated
earlier.)}
The advice to keep track of what has been achieved as you are adding rules
to your program differs from the ``generate-and-test'' advice
in that it refers to mathematical properties of stable models, and not to
programmer's intentions.

\section*{Acknowledgements}

Thanks to Michael Gelfond, Amelia Harrison, Yuliya Lierler, Julian Michael,
Liangkun Zhao, and the anonymous referees for comments on earlier
versions of this paper. 
Conversations and exchanges of email messages with Mark Denecker,
Esra Erdem, Martin Gebser, Roland Kaminski, Johannes Oetsch, Dhananjay Raju,
and Mirek Truszczynski  helped the author develop a
better understanding of the methodology of answer set programming.
This research was partially supported by the National Science Foundation
under Grant IIS-1422455.

\bibliography{bib}
\end{document}